\documentclass{article}
\usepackage{spconf,amsmath,graphicx}
\usepackage{lscape}
\usepackage{xcolor}
\usepackage{algorithm}
\usepackage{algpseudocode}
\usepackage{url}
\usepackage{multirow}
\usepackage{float}
\usepackage{caption}
\usepackage{float}
\usepackage{subfig}
\usepackage{dblfloatfix}
\usepackage{xcolor}
\usepackage{ctable}
\usepackage{times}
\usepackage{epsfig}
\usepackage{graphicx}
\usepackage{amsmath}

\usepackage{amssymb}
\usepackage{subfig}
\usepackage{url}
\usepackage{booktabs}
\usepackage{paralist}
\usepackage{caption}
\renewenvironment{thebibliography}[1]{
	\begin{oldthebibliography}{#1}
		\setlength{\itemsep}{0em}
		\setlength{\parskip}{0em}
	}
	{
	\end{oldthebibliography}
}

\title{Longitudinal Analysis of Mask and No-Mask on Child Face Recognition}
\name{Praveen Kumar Chandaliya\textsuperscript{1}, Zahid Akhtar\textsuperscript{2}, Neeta Nain\textsuperscript{1} \thanks{This work was supported by the MeitY, Government of India, under Grant No. 4 (13)/2019-ITEA''. We are also grateful to Cognitec with the donation of the FaceVACS-DBScanID5.6 software for this work.}}
\address{\textsuperscript{1}Malaviya National Institute of Technology Jaipur, India\\ \textsuperscript{2}State University of New York Polytechnic Institute, USA}

\begin{document}
	%
	\maketitle
	\begin{abstract}
		Face is one of the most widely employed traits for person recognition, even for large-scale applications. Despite technological advancements in face recognition systems (FRS), they still face obstacles caused by pose, expression, occlusion, and aging variations. Owing to the COVID-19 pandemic, contactless identity verification has become exceedingly vital. Recently, few studies have been conducted on the effect of face mask on adult FRS. However, the impact of aging with face mask on child subject recognition has not been adequately explored. Thus, the objective of this study is analyzing the child longitudinal impact together with face mask and other covariates on FRS. Specifically, we performed a comparative investigation of three top performing publicly and a post-COVID-19 commercial-off-the-shelf (COTS) system under child cross-age verification and identification settings using our generated synthetic mask and no-mask samples. Furthermore, we investigated the longitudinal consequence of eyeglasses with mask and no-mask. The study exploited no-mask longitudinal child face dataset (i.e., extended Indian Child Longitudinal Face Dataset) that contains $26,258$ face images of $7,473$ subjects in the age group of $[2, 18]$ over an average time span of $3.35$ years.
		Due to the combined effects of face mask and aging, the FaceNet, PFE, ArcFace, and COTS verification accuracies decrease approximately $25\%$, $22\%$, $18\%$, $12\%$, respectively.
	\end{abstract}

	\section{Introduction}
	\label{sec:introduction}

	Nowadays, face recognition systems under face mask is getting much more momentum. For instance, the work in \cite{Damer2020,NIST6B} evaluated verification performance on both real and synthetic masks. It was later extended in \cite{Damer2021} to analyze the human experts and automatic recognition systems on unmasked, real masked, and synthetic mask on adult dataset. 
	\begin{figure}[t!]%
		\centerline{\includegraphics[width=85mm,height=23mm]{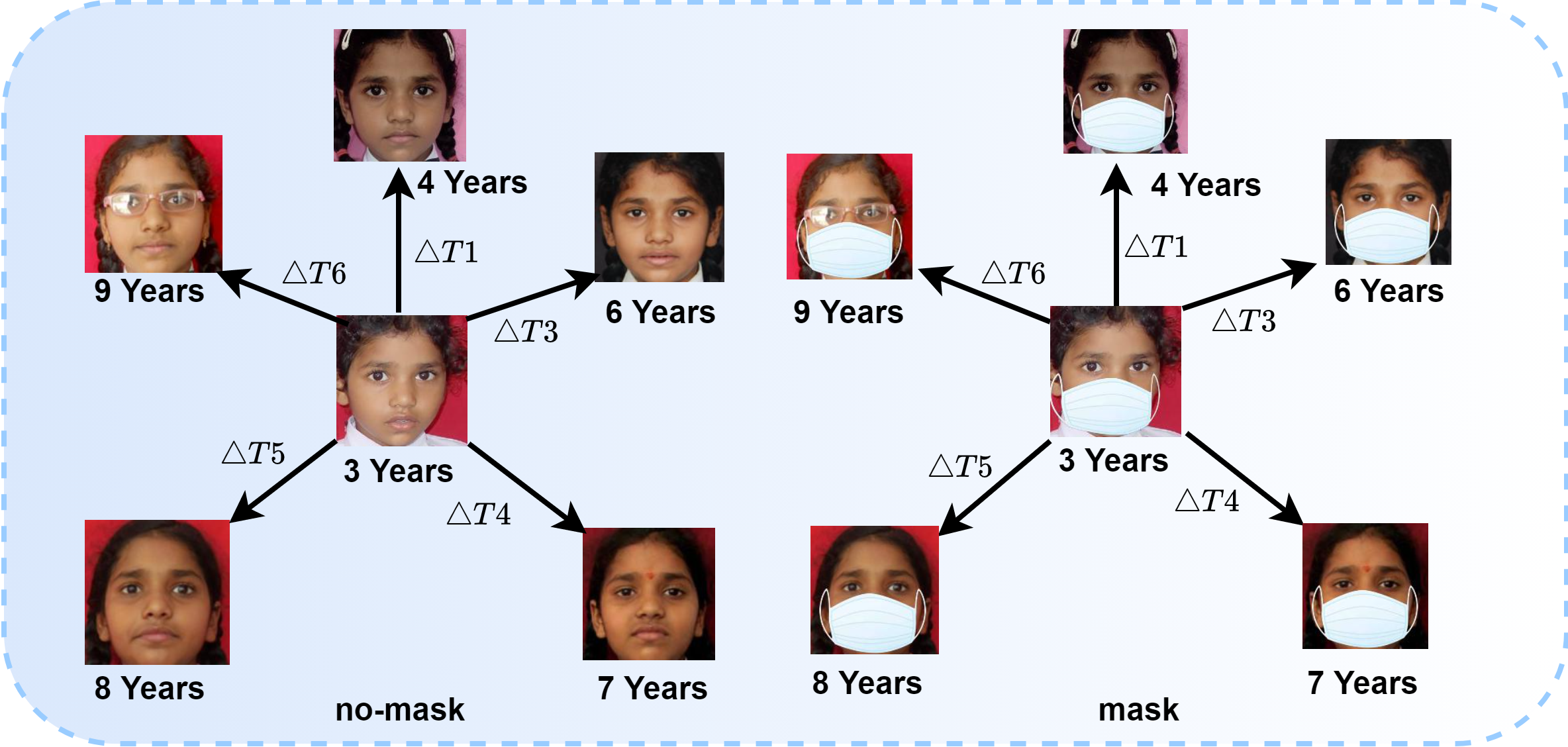}}
		\caption{Left: no-mask subject. Right: mask subject. Center image represents a enrollment image and branches are images of same subject at different ages. Here, T1, T2, T3, T4, T5, and T6 denote time lapses between enrollment and subsequent acquired images. Age at the time of image acquisition (in years) is given below each images.}
		\label{fig:clfmask}%
		\vspace{-18pt}
	\end{figure}%
	\noindent
	\vspace{-3pt}
	
	Besides face mask, face aging is also a vital co-variate that negatively affect automated face recognition systems, especially for child subjects. For example, Deb et al. \cite{Deb2018} fused COTS and FaceNet~\cite{facenet} scores, and attained $80.56\%$ and $53.33\%$ verification accuracy, respectively, for a time lapse of $1$ and $3$ years between enrollment and probe images for subjects of age $[2{-}18]$ years old. The work in \cite{Srinivas2019} investigated five top performing COTS matchers, two government matchers and one open-source face recognition system on Wild Child Celebrity and LFW~\cite{LFWTech} datasets, and obtained maximum of $78.20\%$ and $85.2\%$, respectively, verification and Rank-1 identification accuracy. The study also showed each algorithm's negative bias towards children compared to adult face samples.
	\noindent
	\begin{table}[b!]
		\caption{Number of genuine pairs according to time lapses.}
		\label{tab:tablestatistic}
		\vspace{-8pt}
		\begin{center}
			\scriptsize\addtolength{\tabcolsep}{-0.9pt}
			\begin{tabular}{|c|c|c|c|c|c|c|c|}
				\hline
				&  \multicolumn{6}{c}{\textbf{Genuine pair}}& \\
				\hline
				\textbf{Protocol} & \textbf{Gender} & $\triangle$ T1 & $\triangle$ T2 & $\triangle$ T3 & $\triangle$ T4 & $\triangle$ T5 & $\triangle$ T6  \\
				\hline
				\multirow{2}{*}{\textbf{No-mask}}&  Boys & $2,528$ & $2,622$ & $2,372$ & $1,470$  &$8,11$  &$3,38$  \\
				\cline{2-8}
				{}& Girls & $2,585$ & $2,249$ & $1,670$ & $1,041$  &$5,52$  &$2,60$ \\
				\hline
				\multirow{2}{*}{\textbf{Mask}}&  Boys & $2,506$ & $2,618$ & $2,355$ & $1,452$  &$8,05$  &$3,33$  \\
				\cline{2-8}
				& Girls & $2,579$ & $2,253$ &  $1,663$  &$1,034$  &$5,46$  & $2,54$ \\
				\hline
			\end{tabular}%
			\vspace{-12pt}
		\end{center}%
	\end{table}%
	\begin{figure*}[ht!]%
		\centerline{\includegraphics[width=175mm,height=16mm]{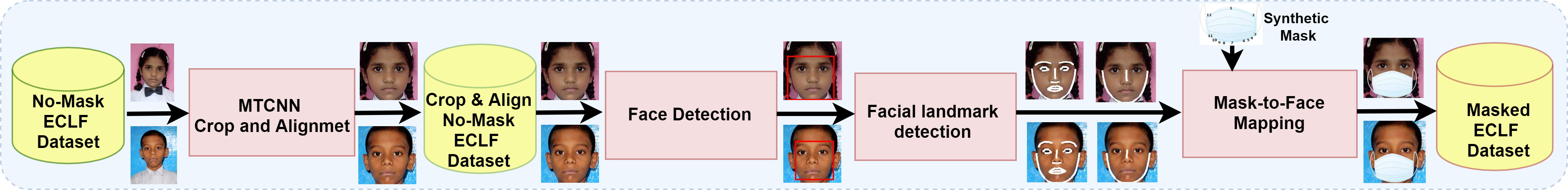}}
		\caption{Pipeline for generating face mask dataset.}
		\label{fig:pipline}%
		\vspace{-10pt}
	\end{figure*}%
	There exist several mask face datasets \cite{MaskedFace2021,Damer2020,Mishra2021,Huang2021} but they mainly contain adult faces and caucasian and north Asian demography. \textit{To the best of our knowledge, no study has explored the combined effect of aging and mask on face recognition when the subjects are children.} Also, no prior works explicitly evaluated the \textit{longitudinal identification} performance when probe samples are with mask and gallery samples are without mask and vice versa. Therefore, this work analyzes the practical covariates (e.g., elapsed time, age, sex, with mask and no-mask, and eyeglasses with mask and no-mask). Namely, we present a longitudinal study using one of the largest, deepest, and longest (in terms of number of subjects, number of images per subject, and time spans of subject images) child face dataset. Thus, this study investigated the above-mentioned directions by extending Children Longitudinal Face (CLF)~\cite{Deb2018,Praveen2021} dataset. We simulated the synthetic mask over all face images by using open source tool Masked Face-Net~\cite{MaskedFace2021} on the children dataset while keeping faces’ longitudinal nature, as also shown in Fig.\ref{fig:clfmask}. 
	There are several venues where child face recognition systems are needed, e.g., finding missing children \cite{Deb2018,Deb2021}, de-duplication of identification documents (e.g., minors passport validation and diver license)~\cite{Lacey2018,Praveen2019} and school attendance during COVID-19 pandemic with mandatory mask \cite{Manfred2020,Zeng2021ASO}. The contributions of this study are as follows:
	\begin{compactitem}[$\bullet$]
		\item A longitudinal child dataset with face samples with synthetic masks.
		\item We conduct extensive \textit{longitudinal} performance analyses of three top-performing public and one COTS face recognition systems on face images of children with mask. There is no such longitudinal study of children to our knowledge.
		\item We provide an extensive comparative evaluation of child longitudinal face verification and identification under joint and disjoint gender with and without face mask. Additionally, we analyze gender bias with eyeglasses and masks.
		\item FRS performance degrades with increasing time between gallery and probe samples. Using a face mask together with aging causes such declines. 
	\end{compactitem}
	The manuscript is organized as follows. Section~\ref{sec:dataset} details the extended longitudinal child dataset used in this study. Section~\ref{sec:experimentsetup} presents the experimental setup. Section~\ref{sec:experimentalanalysis} discusses experimental results. Finally, the conclusion and future works are presented in Section~\ref{sec:conclusion}.
	\section{Longitudinal Children Face Mask Dataset}\label{sec:dataset}
	All publicly available datasets have age range $[14{-}80]$. But there are no publicly available longitudinal masked face recognition datasets specific to children. Therefore, we used extended Children Longitudinal Face (ECLF) dataset that contains $26,258$ face images of $7,473$ subjects in the age group of $[2,18]$. The average number of images per subject is $3$, which were acquired an average over time lapse of $3.35$ years. The ECLF dataset is comprised of $14,057$ $(53.53\%)$ boys and $12,201$ $(46.48\%)$ girls. Statistics for the dataset are shown in Table~\ref{tab:tablestatistic}.

	In Fig.~\ref{fig:pipline} shows main steps of MaskedFace dataset generation. It is worth noting that some faces of ECLF dataset were not able to be processed ($1,550$ boys’ and $55$ girls’ images) because of larger pose and illumination. Hence, the resulting MaskedFace-ECLF contains $24,653$ masked face images ($12,507$ boys and $12,146$ girls) of $7,457$ subjects ($3,732$ boys, $3,725$ girls).

	\begin{figure}[t!]%
		\centerline{\includegraphics[width=90mm,height=42mm]{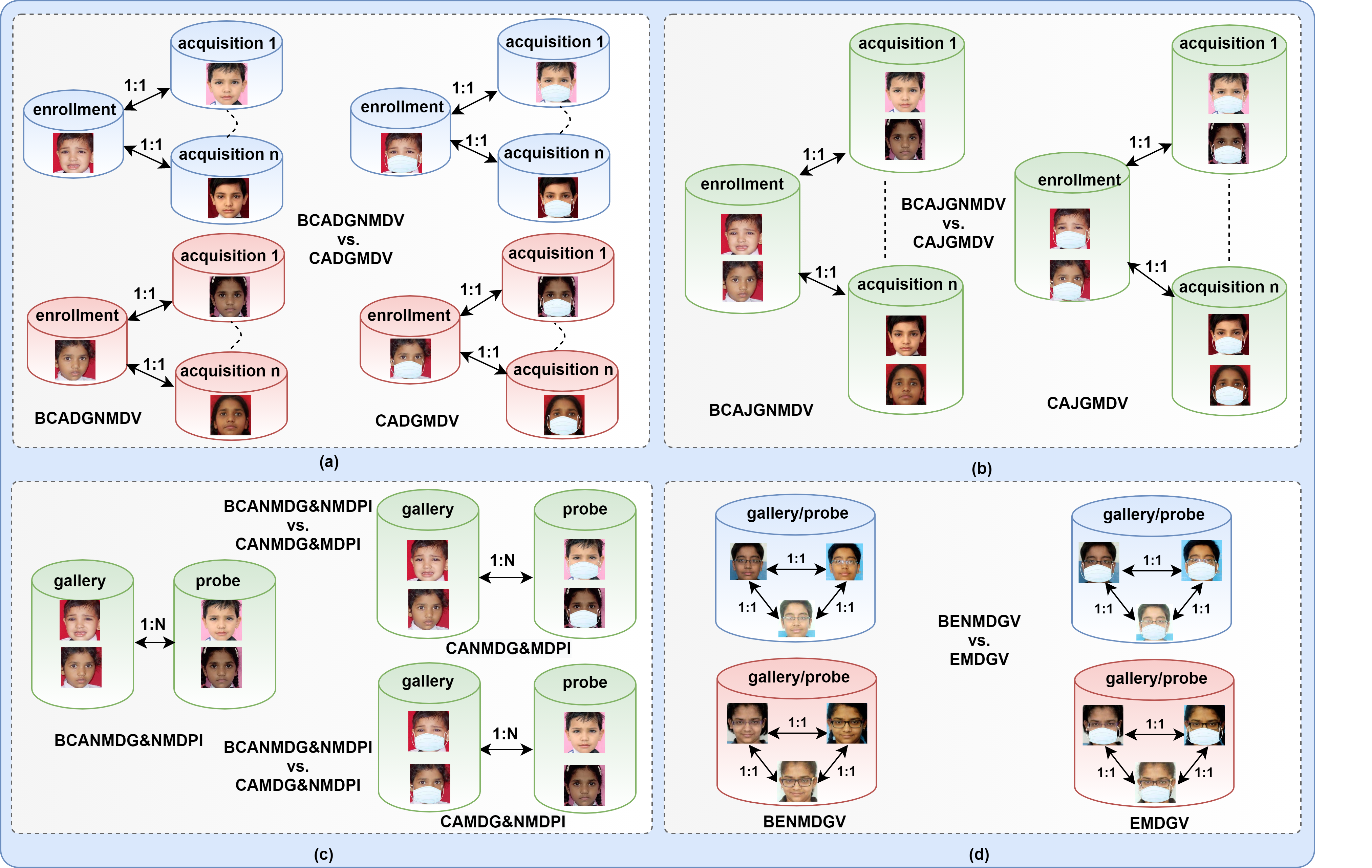}}
		\caption{Four protocols of child longitudinal study with and without mask (zoom for better view). 
		}
		\label{fig:experiment1}%
		\vspace{-16pt}
	\end{figure}
	\section{EXPERIMENTS SETUP}\label{sec:experimentsetup}
	To analyze real-world scenarios of child longitudinal study with mask and no-mask, following four face recognition protocols were investigated, as shown in Fig.\ref{fig:experiment1}.
	\begin{table*}[t!]
		\setlength{\arrayrulewidth}{0.1pt}
		\caption{Longitudinal verification rate $(\%)$ of considered face recognition systems on disjoint gender without mask.}
		\label{tab:disjointwithoutmask}
		\vspace{-8pt}
		\begin{center}
			\scriptsize\addtolength{\tabcolsep}{0.6pt}
			\begin{tabular}{|c|c|c|c|c|c|c|c|c||c|c|c|c|c|c|c|c|}
				\hline
				&  \multicolumn{8}{c||}{\textbf{Boys}} & \multicolumn{6}{c}{\textbf{Girls}} & \\
				\hline
				{\textbf{Model}} & {\textbf{FAR}} & {{$\triangle$T1}} & {{$\triangle$T2}} & {{$\triangle$T3}} & {{$\triangle$T4}} & {{$\triangle$T5}} & {{$\triangle$T6}} & {{Avg}}& {{$\triangle$T1}} & {{$\triangle$T2}} & {{$\triangle$T3}} & {{$\triangle$T4}} & {{$\triangle$T5}} & {{$\triangle$T6}} &{\textbf{Avg}} \\
				\hline
				\multirow{2}{*}{\textbf{FaceNet}} & $1e{-3}$ & 85.99 & 76.08 & 64.12 & 50.13 & 37.60 & 31.95 &\textbf{57.64}& 88.74 & 83.90 & 77.84 & 68.78 & 52.71 & 52.69 &\textbf{70.77} \\
				\cline{2-16}
				& $1e{-4}$ & 70.84 & 57.17 & 40.97 & 28.97 & 21.57 & 16.86 & \textbf{39.40} &75.39&69.71&59.10& 44.76& 30.61& 22.30 & \textbf{50.31}\\
				\hline
				
				\multirow{2}{*}{\textbf{PFE}}& $1e{-3}$ & 99.80 & 99.58 & 98.48 & 94.55 & 92.60 & 86.39 &\textbf{95.23} & 99.69 & 99.33 & 99.28 & 98.94 & 96.55 & 94.23 & \textbf{98.00}\\
				\cline{2-16}
				& $1e{-4}$ & 99.36 & 97.86 & 95.02 & 86.32 & 83.23 & 71.59  &\textbf{88.89} & 99.53&99.07&99.04& 97.50& 93.11& 81.92 &\textbf{95.03}\\
				\hline
				
				\multirow{2}{*}{\textbf{ArcFace}}& $1e{-3}$ & 99.84 & 99.69 & 99.53 & 99.25 & 97.65 & 95.56 & \textbf{98.58}& 99.72 & 99.42 & 99.52 & 99.32 & 98.55 & 96.69 &\textbf{98.87} \\
				\cline{2-16}
				& $1e{-4}$ & 99.60 & 99.46 & 99.03 & 97.41 & 95.19 & 92.01 &\textbf{97.12} & 99.69&99.37&99.40& 99.13& 96.55& 91.92 &\textbf{97.68} \\
				\hline
				
				\multirow{2}{*}{\textbf{COTS}}& $1e{-3}$ & 99.88 & 99.80 & 99.62 & 99.52 & 99.52 & 97.92 & \textbf{99.21} & 99.69 & 99.42 & 99.54 & 99.13 & 97.90 & 96.51 & \textbf{98.69}\\
				\cline{2-16}
				& $1e{-4}$ & 99.72 & 98.20 & 99.49 & 98.29 & 95.68 & 95.26 & \textbf{97.77}&  99.65 &99.24&99.07& 99.27& 96.85& 93.79 &  \textbf{97.81} \\
				\hline
				\hline
				
				{\textbf{Avg}}&  $1e{-4}$ & \textbf{92.38} & \textbf{88.17} & \textbf{83.62} & \textbf{77.75} & \textbf{73.91}&\textbf{68.93}   & \textbf{80.79} &\textbf{93.57}  & \textbf{91.85} & \textbf{89.15} & \textbf{84.92} & \textbf{79.28} & \textbf{72.48} & \textbf{85.21}\\
				\cline{2-16}
				
				\hline
			\end{tabular}%
			\vspace{-18pt}
		\end{center}%
	\end{table*}%
	\begin{table*}[t!]
		\setlength{\arrayrulewidth}{0.1pt}
		\caption{Longitudinal verification rate $(\%)$ of considered face recognition systems on disjoint gender with mask.}
		\label{tab:disjointwithmask}
		\vspace{-8pt}
		\begin{center}
			\scriptsize\addtolength{\tabcolsep}{0.6pt}
			\begin{tabular}{|c|c|c|c|c|c|c|c|c||c|c|c|c|c|c|c|}
				\hline
				&  \multicolumn{8}{c||}{\textbf{Boys}} & \multicolumn{6}{c}{\textbf{Girls}} & \\
				\hline
				{\textbf{Model}} & {\textbf{FAR}} & {{$\triangle$T1}} & {{$\triangle$T2}} & {{$\triangle$T3}} & {{$\triangle$T4}} & {{$\triangle$T5}} & {{$\triangle$T6}} & {\textbf{Avg}}& {{$\triangle$T1}} & {{$\triangle$T2}} & {{$\triangle$T3}} & {{$\triangle$T4}} & {{$\triangle$T5}} & {{$\triangle$T6}} & {{Avg}}\\
				\hline
				\multirow{2}{*}{\textbf{FaceNet}}& $1e{-3}$ & 57.14 & 44.65 & 30.72 & 23.41 & 16.52 &13.81 &\textbf{31.04} & 66.53 & 57.83 & 50.51 & 43.23 & 31.31 & 27.95&\textbf{46.23} \\
				\cline{2-16}
				& $1e{-4}$ & 35.39 & 24.25 & 13.24 & 9.02 & 5.96 & 5.7&\textbf{15.59} &46.62&39.10&24.13& 22.14& 14.46& 8.26&\textbf{25.79}\\
				\hline
				
				\multirow{2}{*}{\textbf{PFE}}& $1e{-3}$ & 94.37 & 89.38 & 80.47 & 69.83 & 59.50 & 52.85 &\textbf{74.40} & 96.77 & 93.90 & 92.12 & 88.78 & 77.83 & 74.80& \textbf{87.37} \\
				\cline{2-16}
				& $1e{-4}$ & 87.15 & 78.53 & 64.52 & 49.17 & 42.23 & 35.73 & \textbf{59.55}& 90.99&87.06&83.34& 74.37& 60.62& 55.90 & \textbf{75.38}\\
				\hline
				
				\multirow{2}{*}{\textbf{ArcFace}} & $1e{-3}$ &97.32  &94.72  & 89.57 & 83.74 &75.40  &67.86 & \textbf{84.76} & 98.82 & 97.46 & 96.51 & 95.26 & 87.91 &83.46 & \textbf{93.24} \\
				\cline{2-16}
				& $1e{-4}$ & 93.25 & 87.66&78.71& 65.28&59.62 &49.55 & \textbf{72.35} & 96.12  & 94.04 & 91.70 & 85.78 & 72.89 & 67.71 & \textbf{84.71} \\
				\hline
				
				\multirow{2}{*}{\textbf{COTS}} & $1e{-3}$ & 97.95 & 96.01 & 91.87 & 88.07 & 86.88 & 85.54 & \textbf{91.05} &  97.74 & 98.16 & 97.58 & 96.50 & 93.21& 92.88 & \textbf{96.01} \\
				\cline{2-16}
				& $1e{-4}$ & 94.42 & 90.88 & 82.92 & 73.64 & 73.07 & 72.89 & \textbf{81.30} &95.41&95.89& 94.01& 91.26& 89.46& 86.95& \textbf{92.16} \\
				\hline
				\hline
				{\textbf{Avg}}&  $1e{-4}$ & \textbf{77.55} & \textbf{70.33} & \textbf{59.85} & \textbf{49.28} & \textbf{45.22}&\textbf{40.97} &\textbf{57.20}   & \textbf{82.28} &\textbf{79.02}  & \textbf{73.30} & \textbf{68.39} & \textbf{59.35} & \textbf{54.71} & \textbf{69.51} \\
				\hline
			\end{tabular}
			\vspace{-18pt}
		\end{center}
	\end{table*}
	
	\noindent\textbf{BCADGNMDV \textnormal{vs.} CADGMDV:} This protocol evaluates cross-age face verification performance under no-mask and mask with disjoint gender influence. It is done by performing $1{:}1$ comparison, where enrollment image (first acquired image at youngest age) is compared to subsequent face images of the same subject at greater age than enrollment. We named this protocol \textbf{B}aseline \textbf{C}ross-\textbf{A}ge \textbf{D}isjoint \textbf{G}ender \textbf{N}o-\textbf{M}aske\textbf{D} \textbf{V}erification vs. \textbf{C}ross-\textbf{A}ge \textbf{D}isjoint \textbf{G}ender \textbf{M}aske\textbf{D} \textbf{V}erification (BCADGNMDV \textnormal{vs.} CADGMDV).
	
	\noindent\textbf{BCAJGNMDV \textnormal{vs.} CAJGMDV:} To compare joint effect of gender and aging with mask and no-mask, we used same $1{:}1$ cross-age verification strategy as in (BCADGNMDV \textnormal{vs.} CADGMDV) but with joint gender. We named this cross-age protocol as \textbf{B}aseline \textbf{C}ross-\textbf{A}ge \textbf{J}oint \textbf{G}ender \textbf{N}o-\textbf{M}aske\textbf{D} \textbf{V}erification vs. \textbf{C}ross-\textbf{A}ge \textbf{J}oint \textbf{G}ender \textbf{M}aske\textbf{D} \textbf{V}erification (BCAJGNMDV  \textnormal{vs.} CAJGMDV).

	\noindent\textbf{BCANMDG\&NMDPI \textnormal{vs.} CANMDG\&MDPI \textnormal{and} BCANMDG\&NMDPI \textnormal{vs.} CAMDG\&NMDPI:} This protocol simulates two real time identification cases: (i) time of reopening school and (ii) missing child recognition in pandemic. In former case, the gallery set faces were with no-mask and the probe set faces were with mask. We perform cross-age identification comparison between all gallery enrollment samples of all subjects and probe non-enrollment samples. Particularly, we conduct joint gender $1{:}N$ close-set \textbf{B}aseline \textbf{C}ross-\textbf{A}ge \textbf{N}o-\textbf{M}aske\textbf{D} \textbf{G}allery and \textbf{N}o-\textbf{M}aske\textbf{D} \textbf{P}robe \textbf{I}dentification vs. \textbf{C}ross-\textbf{A}ge \textbf{N}o-\textbf{M}aske\textbf{D} \textbf{G}allery and \textbf{M}aske\textbf{D} \textbf{P}robe \textbf{I}dentification (BCANMDG\&NMDPI \textnormal{vs.} CANMDG\&MDPI). In latter case, the gallery set faces were with mask and the probe set faces were with no-mask. We conduct joint gender $1{:}N$ close-set \textbf{B}aseline \textbf{C}ross-\textbf{A}ge \textbf{N}o-\textbf{M}aske\textbf{D} \textbf{G}allery and \textbf{N}o-\textbf{M}aske\textbf{D} \textbf{P}robe \textbf{I}dentification vs. \textbf{C}ross-\textbf{A}ge \textbf{M}aske\textbf{D} \textbf{G}allery and \textbf{N}o-\textbf{M}aske\textbf{D} \textbf{P}robe \textbf{I}dentification (BCANMDG\&NMDPI \textnormal{vs.} CAMDG\&NMDPI).
	
	\noindent\textbf{BENMDGV \textnormal{vs.} EMDGV:} This protocol analyzes the effect of eyeglass with no-mask and mask face longitudinal verification performance. We perform $1{:}1$ comparison between \textbf{B}aseline  \textbf{E}yeglass \textbf{N}o-\textbf{M}aske\textbf{D}  \textbf{D}isjoint \textbf{G}ender \textbf{V}erification vs. \textbf{E}yeglass \textbf{M}aske\textbf{D} \textbf{D}isjoint \textbf{G}ender \textbf{V}erification (BENMDGV \textnormal{vs.} EMDGV). 
	\section{EXPERIMENTAL RESULTS}
	\label{sec:experimentalanalysis}
	\vspace{-7pt}
	We present the cross-age verification, cross-age identification, and verification performance achieved by the four FRS. 
	\vspace{-9pt}
	\subsection{BCADGNMDV \textnormal{vs.} CADGMDV}
	In Table \ref{tab:disjointwithoutmask}, we report results of longitudinal verification rate of face recognition systems with gender disjoint and without mask. Several observations can be obtained from Table \ref{tab:disjointwithoutmask}. For instance, at $0.1\%$ FAR operating point, the average accuracy over $\triangle\textnormal{T1}$ to $\triangle\textnormal{T6}$ for boys ranges from $57.64\%$ (by FaceNet) to $99.21\%$ (by COTS).
	\begin{table*}[t!]
		\setlength{\arrayrulewidth}{0.08pt}
		\caption{Longitudinal verification rate (\%) of face recognition systems on joint gender with and without mask.}
		\label{tab:table_boygirlwomask}
		\vspace{-8pt}
		\begin{center}
			\scriptsize\addtolength{\tabcolsep}{0.6pt}
			\begin{tabular}{|c|c|c|c|c|c|c|c|c||c|c|c|c|c|c|c|}
				\hline
				{\textbf{Protocol}}&  \multicolumn{8}{c||}{\textbf{Boys and Girls with No-Mask}}&   \multicolumn{6}{c}{\textbf{Boys and Girls with Mask}}& \\
				\hline
				{\textbf{Model}} & {\textbf{FAR}} & {{$\triangle$T1}} & {{$\triangle$T2}} & {{$\triangle$T3}} & {{$\triangle$T4}} & {{$\triangle$T5}} & {{$\triangle$T6}}& {{Avg}} & {{$\triangle$T1}} & {{$\triangle$T2}} & {{$\triangle$T3}} & {{$\triangle$T4}} & {{$\triangle$T5}} & {{$\triangle$T6}}& {{Avg}} \\
				
				\hline
				\multirow{2}{*}{\textbf{FaceNet}}& $1e{-}3$ & 89.57 & 82.03 & 73.33 & 61.69 & 46.73 & 43.31 & \textbf{66.11} & 65.19& 55.20 & 43.40 & 34.79& 23.61 & 23.50  & \textbf{40.95}  \\
				\cline{2-16}
				& $1e{-}4$ & 77.08 & 65.55 & 52.20 & 39.54 & 26.85 & 25.25 & \textbf{47.75} & 44.56 & 33.05 & 23.02 & 16.85 & 10.51 & 10.22 & \textbf{23.04} \\
				\hline
				\multirow{2}{*}{\textbf{PFE}}& $1e{-}3$ & 99.76 & 99.55 & 99.03 & 97.53 & 94.64 & 90.96 &  \textbf{96.91} & 96.30 & 93.06  & 87.33 & 80.25 & 68.39 & 64.22  & \textbf{81.59}\\
				\cline{2-16}
				& $1e{-}4$ & 99.59 & 98.70 & 97.37 & 92.47 & 88.48 & 78.76 & \textbf{92.56} & 90.37 & 84.70 & 74.91 & 62.83 & 58.18 & 46.67 & \textbf{69.61}  \\
				\hline
				\multirow{2}{*}{\textbf{ArcFace}} & $1e{-}3$ &99.78  &99.60  & 99.57 &99.32  &  98.16  & 96.48 & \textbf{98.82} &98.26  &96.75  & 93.72 &90.38  &  82.01  & 77.51 & \textbf{89.77} \\
				\cline{2-16}
				& $1e{-}4$ & 99.72 & 99.50 & 99.25 &98.56  &96.47  &94.14 &  \textbf{97.94} & 95.24 & 91.72 & 85.39 &78.64  &66.69  &59.79 & \textbf{79.58} \\
				\hline
				
				\multirow{3}{*}{\textbf{COTS}}& $1e{-}3$ & 99.86 &  99.73 & 99.65 & 99.32 & 97.94 & 97.48 & \textbf{98.99} & 98.63 &  97.55 & 95.44 & 92.78 & 88.92 & 88.44 & \textbf{93.63} \\
				\cline{2-16}
				& $1e{-}4$ & 99.69 & 99.44  & 99.00  & 97.80 & 93.08 & 92.60  & \textbf{96.94} & 90.11 & 93.34  & 90.08  & 84.75 & 78.44 & 77.13 & \textbf{85.64} \\
				\hline					
				\hline
				{\textbf{Avg}}&  $1e{-}4$ &\textbf{94.02}   &  \textbf{90.80} &\textbf{86.96} & \textbf{82.10} &\textbf{76.22}  &  \textbf{72.69} & \textbf{83.79} &\textbf{80.07}   &  \textbf{75.70} &\textbf{68.35} & \textbf{60.77} &\textbf{53.46}  &  \textbf{48.45} & \textbf{64.47} \\
				\hline
			\end{tabular}%
			\vspace{-15pt}
		\end{center}%
	\end{table*}%
	
	Whereas, it ranges from $70.77\%$ (by FaceNet) to $98.69\%$ (by COTS) for girls. As the age time lapse between gallery and probe samples increases, the accuracy of face systems decreases, e.g., the COTS verification rates with $0.01\%$ FAR operating point for boys are $99.72\%$ at $\triangle\textnormal{T1}$ and $95.26\%$ at $\triangle\textnormal{T6}$. Based on majority voting, we can state that all considered face systems achieved better performances for girls than boys under all-time lapses. For example, the average accuracies with $0.01\%$ FAR operating point of boys and girls at $\triangle\textnormal{T3}$ are $83.62\%$ and $89.15\%$, respectively. Similar face systems’ bias towards girls/females was reported in \cite{Deb2018}.
	\begin{table*}[ht!]
		\setlength{\arrayrulewidth}{0.01pt}
		\caption{Longitudinal closed-set identification rate (\%) of joint gender face recognition systems on (gallery vs. probe) no-mask vs. no-mask, no-mask vs. mask and mask vs. no-mask.}
		\label{table_example}
		\vspace{-9pt}
		\begin{center}
			\tiny\addtolength{\tabcolsep}{-1.1pt}
			\begin{tabular}{|c|c|c|c|c|c|c|c|c||c|c|c|c|c|c|c||c|c|c|c|c|c|c|}
				\hline
				{\textbf{Protocol}}&  \multicolumn{8}{c||}{\textbf{No-mask vs. No-mask}} & \multicolumn{6}{c}{\textbf{No-mask vs. Mask}} & & \multicolumn{6}{c}{\textbf{Mask vs. No-mask}} &  \\
				\hline
				{\textbf{Model}} & {\textbf{Rank}} & {{$\triangle$T1}} & {{$\triangle$T2}} & {{$\triangle$T3}} & {{$\triangle$T4}} & {{$\triangle$T5}} & {{$\triangle$T6}} & {\textbf{Avg}} & {{$\triangle$T1}} & {{$\triangle$T2}} & {{$\triangle$T3}} & {{$\triangle$T4}} & {{$\triangle$T5}} & {{$\triangle$T6}} & {{Avg}} & {{$\triangle$T1}} & {{$\triangle$T2}} & {{$\triangle$T3}} & {{$\triangle$T4}} & {{$\triangle$T5}} & {{$\triangle$T6}} & {\textbf{Avg}} \\
				\hline
				\multirow{1}{*}{\textbf{FaceNet}}&  R-1 & 85.31 & 78.95 & 71.21 & 65.87 &56.73  &61.20 & \textbf{69.68}  & 31.09 & 28.26 & 25.08 & 22.42 &20.86  &28.11 & \textbf{25.97} & 33.72 & 29.28 & 23.25 & 22.71 &20.78  & 24.53 & \textbf{24.11} \\
				\hline
				\multirow{1}{*}{\textbf{PFE}}&   R-1 & 98.88& 99.06 &98.53  &98.07  & 96.32 &93.97  & \textbf{97.47} & 88.86& 83.50 & 81.14  & 77.93 & 75.41  & 76.66 & \textbf{80.58} & 91.50 &88.69 &  85.55 & 80.32 & 78.17   & 79.73 & \textbf{84.85} \\
				\cline{2-9}
				\hline
				\multirow{1}{*}{\textbf{ArcFace}}&   R-1 &99.41 & 99.33 & 99.03  & 98.80& 97.86 & 96.49 & \textbf{98.49} &95.43 & 93.62 & 91.08  & 88.36& 85.47 &87.90 & \textbf{89.98} &92.21 & 90.72  & 85.95  &80.81 &75.78 & 75.64 &\textbf{83.52}  \\
				\hline
				\multirow{1}{*}{\textbf{COTS}}&  R-1  & 98.27 & 99.09  & 98.66& 98.54 & 97.64 & 96.99 & \textbf{98.20} & 90.47& 89.26 & 86.62  & 83.64& 75.86 & 78.53 & \textbf{84.07} &94.06 & 93.89 &   90.57 &86.91 &82.93 & 79.89 &\textbf{88.04} \\
				\hline
				\hline
				
				{\textbf{Avg}}&  R-1  & \textbf{95.48} & \textbf{94.11}  &\textbf{91.86} & \textbf{90.32} & \textbf{87.14} & \textbf{87.16} & \textbf{91.01} & \textbf{76.46} & \textbf{73.66}  & \textbf{70.98} & \textbf{67.59} & \textbf{64.40}  & \textbf{67.80} & \textbf{70.15} & \textbf{77.87} & \textbf{75.65} & \textbf{71.32}   & \textbf{67.69} & \textbf{64.41}  & \textbf{64.95}  &\textbf{70.13}   \\
				\hline
			\end{tabular}%
			\vspace{-18pt}
		\end{center}%
	\end{table*}
	Moreover, we analyzed the skin tones of boys and girls by selecting a $3\times3$ patch from forehead of the subject, then we averaged the patch values as a skin tone indicator. The average skin tone indicator for boys and girls, in the used dataset, is $166.07$ and $176.59$, respectively. Namely, the girls’ skin tones are lighter than boys, and it has been reported in many studies, e.g., \cite{sixta2020fairface}, that face systems attain better performances on lighter skin subjects. Also, we found that more boy subjects are with eyeglasses than girls that may be another variate negatively impacting the face systems. Among FaceNet~\cite{facenet}, PFE~\cite{PFE}, ArcFace~\cite{arcface2018} and COTS face systems~\cite{cognitec}, COTS outperformed others consistently for all time lapses. However, among three academic face systems, FaceNet and ArcFace, respectively, achieved worst and best performances, because FaceNet uses softmax loss function which is known for not being capable of discriminating hard pairs\cite{arcface2018}. Whereas, ArcFace is based on additive angular margin loss that simultaneously enhances the intra-class compactness and inter-class discrepancy. Similar observations can be seen in Table \ref{tab:disjointwithmask} for CADGMDV experiment. Besides, we can notice in Tables \ref{tab:disjointwithoutmask} and \ref{tab:disjointwithmask} that face mask decreases the performances of face systems. For example, the average verification rates at $0.1\%$ FAR operating point using PFE for girls without and with mask, respectively, are $98.0\%$ and $87.37\%$. Also, it is evident that face mask with aging leads to a greater performance degradation than only with mask, e.g., for boys without mask, the average verification rate with $0.01\%$ FAR operating point at $\triangle\textnormal{T1}$ is 92.38\%, while it is $40.97\%$ with mask at $\triangle\textnormal{T6}$.
	\vspace{-13pt}
	\noindent
	\subsection{BCAJGNMDV  \textnormal{vs.} CAJGMDV}
	Table~\ref{tab:table_boygirlwomask} shows the results of joint gender when both probe and gallery samples are without mask (`Boys and Girls with No-Mask’) and when both probe and gallery samples are with mask (`Boys and Girls with Mask’). It can be seen in Table~\ref{tab:table_boygirlwomask} that the performances of the systems are optimal when the acquisition time delay between probe and gallery is small (i.e., time lapse T1). Also, the face systems attained lower cross-age verification performance when both probe and gallery samples are with mask than when both probe and gallery samples are without mask.
	\vspace{-8pt}
	\begin{table}[hb!]
		\caption{Verification rate (\%) of disjoint gender face systems on no-mask with eyeglasses and mask with eyeglasses.}
		\label{tab:table_spectacle}
		\vspace{-11pt}
		\begin{center}
			\scriptsize\addtolength{\tabcolsep}{-5.5pt}
			\begin{tabular}{|c|c||c|c||c|c||c|c||c|c||c|c|c|c||c|}
				\hline
				{\textbf{Protocol}} & {\textbf{Gender}} &  \multicolumn{2}{c||}{\textbf{FaceNet}}& \multicolumn{2}{c||}{\textbf{PFE}}&  \multicolumn{2}{c||}{\textbf{ArcFace}}&   \multicolumn{2}{c||}{\textbf{COTS}} & {\textbf{Avg}} \\
				\hline
				{} & {\textbf{FAR}} &  {{$1e{-}3$}} & {{$1e{-}4$}} &  {{$1e{-}3$}} & {{$1e{-}4$}} & {{$1e{-}3$}} & {{$1e{-}4$}}  & {{$1e{-}3$}} & {{$1e{-}4$}} &{{$1e{-}4$}}  \\
				\hline
				\multirow{2}{*}{\textbf{No-mask+}}&  Boys  & 	68.70  & 48.80 &  99.34  & 96.55 &  100.0 & 99.73  &  99.86 & 99.60 & \textbf{86.17} \\
				\cline{2-11}
				{\textbf{Eyeglass}}& Girls & 81.83 & 66.57  &99.47 & 98.54  & 100.0 & 99.86  &100.0 & 100.0 & \textbf{91.24}\\
				\hline
				\multirow{2}{*}{\textbf{Mask+}}&  Boys  & 27.29 & 11.85   &78.02  &58.45& 85.08 & 69.24  & 90.45 & 81.28 &  \textbf{55.21}   \\
				\cline{2-11}
				\textbf{Eyeglass}& Girls  &49.07&  27.58    &89.65 & 77.58 & 96.15  & 87.93  & 97.47 & 94.42 & \textbf{71.88}  \\
				\hline
			\end{tabular}%
			\vspace{-17pt}
		\end{center}
	\end{table}
	\vspace{-14pt}
	\subsection{BCANMDG\&NMDPI \textnormal{vs.} CANMDG\&MDPI \textnormal{and} BCANMDG\&NMDPI \textnormal{vs.} CAMDG\&NMDPI}
	\vspace{-4pt}
	In Table \ref{table_example}, we report the results of longitudinal closed-set identification rate of face systems on no-mask vs. no-mask, no-mask vs. mask and mask vs. no-mask scenarios. The mask vs. no-mask is representation of gallery vs. probe situation, where a missing child’s photo is only with mask (i.e., gallery sample) and after some years it is being compared with no-mask probe sample. We can see in the Table \ref{table_example} that, like in verification, the identification’s performance decreases with increase in time lapses, but the rate of degradation is smaller. Moreover, the identification accuracy suffers from mask as well as compound effect of mask and aging, e.g., FaceNet at $\triangle\textnormal{T1}$ attained $85.31\%$ Rank-1 accuracy in no-mask vs. no-mask, whereas it reached $28.11\%$ at $\triangle\textnormal{T6}$ in no-mask vs. mask. In Table \ref{table_example}, in both no-mask vs. mask and mask vs. no-mask settings, performances varies based on the face algorithms. Moreover, the results show that on an average the mask vs. no-mask scenario achieved better accuracy than no-mask vs. mask setting. All in all, ArcFace and COTS achieved comparative performances in identification mode.
	\vspace{-13pt}
	\noindent
	\subsection{BENMDGV \textnormal{vs.} EMDGV}
	\vspace{-2pt}
	The objective of this experiment is to study gender bias with eyeglasses and mask on verification. Out of $26,258$ images in ECLF dataset, only $1,718$ images from $396$ boy and $346$ girl subjects are with eyeglasses. For fairness, we selected $2,22$ subjects for each boy and girl group, where $88$ subjects are with $2$ images, $70$ with $3$ images, $46$ with $4$ images and $18$ with $5$ images. We can observe in Table~\ref{tab:table_spectacle} that even though boys and girls subjects are with eyeglasses but for girls the FRS achieved higher performances in both no-mask+eyeglass (i.e., both template and query are without mask) and mask+eyeglass (i.e., both template and query are with mask) setting. For example, COTS procured $99.60\%$ and $100\%$ (for no-mask) and $81.28\%$ and $94.42\%$ (for mask) at FAR $0.01\%$ for boys and girls, respectively. It is also easy to see that mask+eyeglasses lessen the accuracies of the FRS, e.g., FaceNet accuracy diminished from $68.70\%$ to $27.29\%$ for boys. For eyeglass+(no-) mask, COTS performed better than ArcFace.
	\vspace{-12pt}
	\section{Conclusion and Future Work}\label{sec:conclusion}
	Driven by the COVID-19 pandemic and subsequent face mask conformity, this paper, contrary to prior works, investigated the impact of aging with face mask on child subject recognition. Particularly, the empirical efficacy of four FRS was conducted under face masked children cross-age verification and identification scenarios. This study assembled longitudinal Indian children (i.e., boys and girls aged from 2 to 18) cohorts database with synthetic masks, and showed that face systems’ performances are severely deteriorated by aging with masks. Moreover, the study found that accuracy of FRS is affected by mask with eyeglasses. Also, the identification levels of girls in the ECLF appear to be higher than boys. In future, we will work towards creating a longitudinal child database with real masks and different ethnicities, developing FRS that are inherently robust to face mask aging, and investigating face mask aging as a face presentation attack.

	

	\bibliographystyle{IEEEbib}
	\bibliography{refs}
	
\end{document}